# Multimodal Federated Learning in Healthcare: a Review

Jacob Thrasher[1*], Alina Devkota[1], Prasiddha Siwakotai[2], Rohit Chivukula[1], Pranav Poudel[2], Chuanbo Hu[3], Ahmad Tafti[4], Binod Bhattarai[5], Prashnna Gyawali[1*]

[1]Lane Department of Computer Science and Electrical Engineering, West Virginia University, Morgantown, WV, USA
[2]Instutiute of Engineering, Tribhuvan University, Kathmandu, Nepal
[3]Department of Computer Science, University at Albany, Albany, NY, USA
[4]Department of Health Information Management, University of Pittsburgh, Pittsburgh, USA
[5]School of Natural and Computing Sciences, University of Aberdeen, Aberdeen, Scotland

*Co-corresponding authors: jdt0025@mix.wvu.edu, prashnna.gyawali@mail.wvu.edu;
Contributing authors: ad00139@mix.wvu.edu, prasiddha.siwakoti43@gmail.com, rc00008@mix.wvu.edu, poudelpranav@gmail.com, chu3@albany.edu, tafti.ahmad@pitt.edu, binod.bhattarai@abdn.ac.uk

**Abstract:**
Recent advancements in multimodal machine learning have empowered the development of accurate and robust AI systems in the medical domain, especially within centralized database systems. Simultaneously, Federated Learning (FL) has progressed, providing a decentralized mechanism where data need not be consolidated, thereby enhancing the privacy and security of sensitive healthcare data. The integration of these two concepts supports the ongoing progress of multimodal learning in healthcare while ensuring the security and privacy of patient records within local data-holding agencies. This paper offers a concise overview of the significance of FL in healthcare and outlines the current state-of-the-art approaches to Multimodal Federated Learning (MMFL) within the healthcare domain. It comprehensively examines the existing challenges in the field, shedding light on the limitations of present models. Finally, the paper outlines potential directions for future advancements in the field, aiming to bridge the gap between cutting-edge AI technology and the imperative need for patient data privacy in healthcare applications.

## Introduction:

Artificial intelligence (AI) tools have been transforming several domains (for example, language translation, speech recognition), and in recent years, it has been showing promise in healthcare applications. Most of such demonstrations have been narrowly focused on tasks using a single modality. However, with advancements in healthcare technology, we now have multiple sources of information available, including medical scans, clinical data, biospecimen data, and several omics data. Collectively, these multiple modes of data could provide crucial information for general understanding of health mechanics, including prognosis, treatment, and prevention.

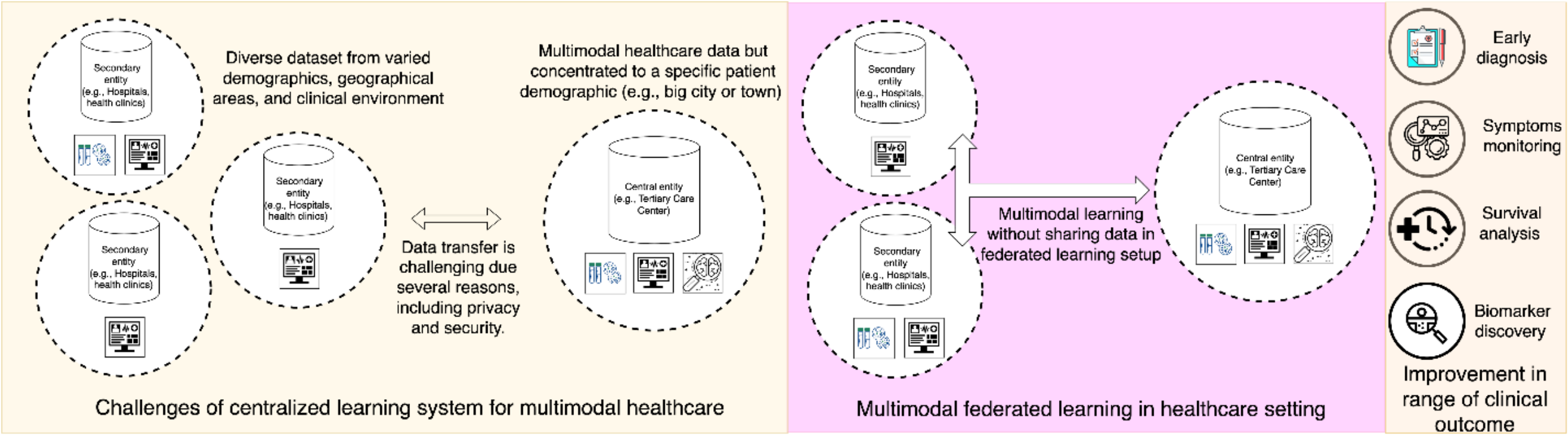

***Figure 1*** *An overview of the Multimodal Federated Learning System. This system is adept at processing multi-modality data from various sources, such as hospitals and health centers, in contrast to centralized learning systems. It aims to improve performance in a wide range of clinical applications, showcasing its versatility and efficiency in healthcare analytics.*

The transformation of AI has been mostly happening with the rise of deep learning which requires many training examples. However, healthcare data, mostly due to privacy reasons, are not easily available for deep learning training. On top of that, these data often present themselves at hospital or health clinics and gathering and collecting at centralized locations to meaningfully process such wealth of data from multiple sources is a significant challenge and requires novel tools and framework. Federated learning (FL) [1] has come out as an efficient mechanism that coordinates the deep learning training through the central server under the premise that the data is stored locally, potentially at different hospitals or clinics. Most of the current applications of federated learning in healthcare have been focused on unimodal data. In contrast, the clinical decision-making process involves analyzing multiple sources and modalities, and rightly, healthcare systems often have such private multiple sources of data available. The development of multimodal federated learning where multiple entities (hospitals) collaborate for one or multiple clinical tasks, under the coordination of a central entity, incorporating data across modalities—including imaging, clinical, genetic, and several healthcare data – is poised to enable a broad range of healthcare applications. We provide the general overview in Figure 1.

In this review, we summarize major advances and challenges in multimodal federated learning in healthcare, providing a concise overview of the current state and future opportunities. We first summarize recent progress and highlight studies that have demonstrated the utility of federated learning in healthcare. We then briefly highlight the importance of multimodal learning in healthcare and inspect promising directions for medical AI research in the form of multimodal federated learning. Finally, we discuss major challenges for successful realization of such multimodal federated systems, including algorithmic limitations of current systems, challenges around data heterogeneity and domain shifts. Details of general AI algorithms for health, including multimodal AI will not be discussed here but are reviewed elsewhere [2], [3], [4].

## Paper Aggregation Methods

Aggregating relevant and reliable studies is vital for any robust review. A quality survey cannot be guaranteed without a proper plan for sourcing articles. Federated learning is a relatively new machine learning paradigm, first emerging in 2016 [5]. Since then, a great deal of research has been conducted to apply this method of distributed training in the healthcare domain. Despite this, there is still a lack of research extending FL in healthcare to multimodal applications as most of the current work focuses on unimodal data. Thus, sourcing relevant articles was a challenge. We began with a breadth-first approach by broadly exploring papers related to "Multimodal Federated Learning in Healthcare". From there, five primary application domains were identified in current trends. All relevant work could be categorized as one of the following tasks: disease detection/diagnosis, segmentation, human activity recognition, survival analysis, and MRI reconstruction. We then proceeded to search each category in depth via "(Multimodal/cross-modal) Federated Learning <task>", where one or none of "(Multimodal/cross-modal)" is selected and "<task>" corresponds to the desired task. Finally, to ensure full coverage of the topic, citation mining was performed within particularly relevant papers. This method yielded a total of 20 papers relevant to multimodal federated learning (MMFL) in healthcare, listed in Table 1. It should be noted that while non-healthcare MMFL works could influence future research in medicine, we believe them to be out of scope as this review primarily focuses on the current, specific applications within healthcare rather than a technical analysis of generalized MMFL.

## Federated Learning in Healthcare: An overview

Artificial Intelligence and Machine Learning (ML) have made great strides in healthcare in recent years. Conventional ML approaches follow a centralized data approach whereby thousands of feature/label pairs are contained within one large database, which is then used to train a model for some predefined tasks such as classification or segmentation. Medical records are one of the most private and protected pieces of information humans have. As such, its distribution is heavily regulated. Health Insurance Portability and Accountability Act (HIPAA) [6] and the General Data Protection Regulation (GDPR) [7] are some examples. These privacy acts make curating and distributing massive datasets in the medical domain difficult, or at times impossible. Besides privacy concerns, centralizing data poses other regulatory, ethical, legal, and technical challenges.

Initially proposed by McMahan et. al in 2016 [1], Federated Learning (FL) is a decentralized approach which aims to address these issues by distributing training across edge devices without compromising privacy [8], [9], [10]. FL can help train models collaboratively while keeping local data private. Under the FL paradigm, each node trains a model using its local data and periodically sends the model parameters to the central server. The central server collects and aggregates the parameters uploaded by all nodes and creates a global model that is sent back to the nodes for further training. During these processes, a node's training data is never shared with other nodes or the central server; only the trainable weights or updates to the model are shared [11]. In healthcare, this could involve distributing models to a collection of semi-centralized data agencies such as hospitals, which use their own data to train a local model before sending the weights to the cloud for aggregation [12], [13]. A client can also refer to an entity in a fully decentralized system where each client only holds data for a single individual such as an IoT device (i.e., Google Home) [14] or personal smartphone [15]. We provide a visualization overview of centralized, semi-centralized, and fully centralized systems in Figure 2. A federated learning system that can efficiently distribute and aggregate knowledge from such a variety of data sources, ranging from centralized hospitals to personalized devices, holds immense potential to unlock the true capabilities of AI in healthcare.

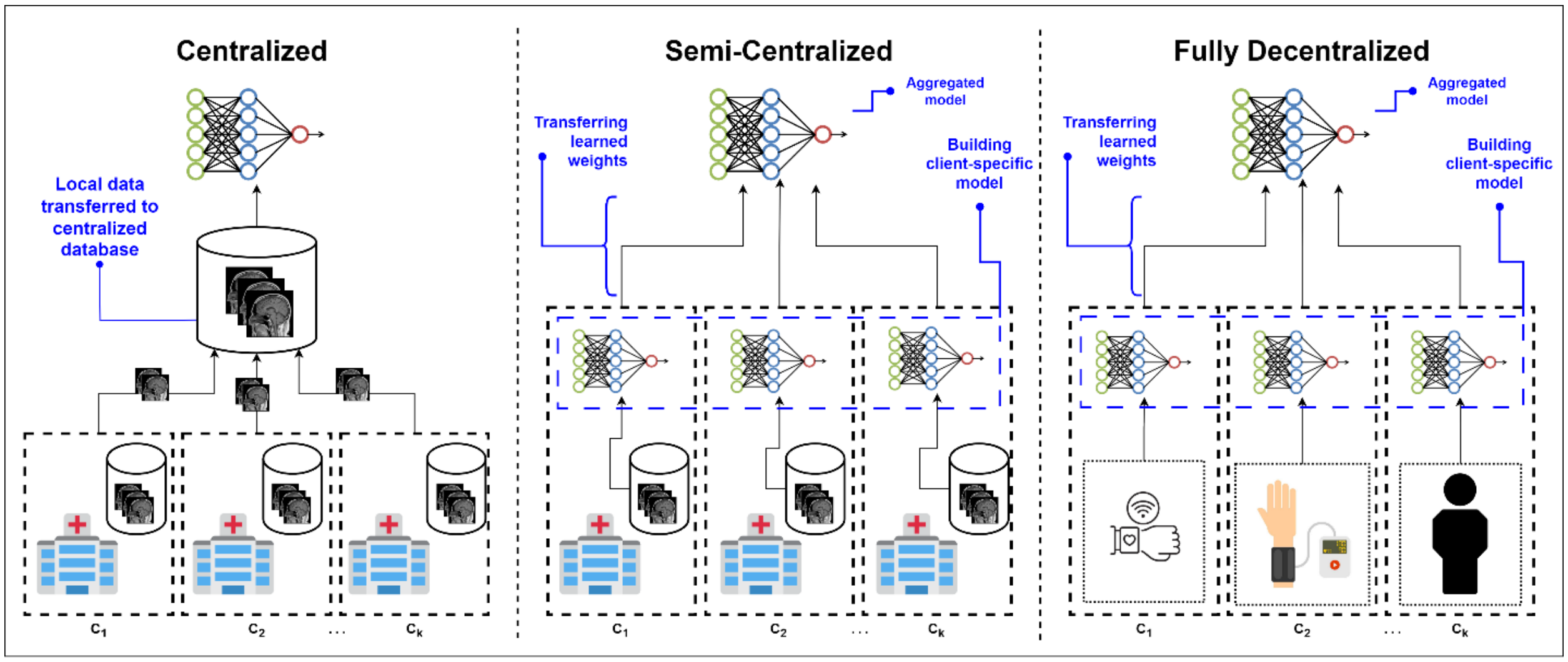

***Figure 2: (Left)*** *Centralized ML, a non-federated learning system characterized by raw data (e.g., medical scans) from multiple data-holding agencies held in a central database for training.* ***(Middle)*** *Semi-centralized Federated Learning, where each client maintains its own private dataset and builds client-specific models, which are trained locally and aggregated in the server.* ***(Right)*** *Fully Decentralized Federated Learning, where rather than data-holding agencies, clients are individuals who contain their own personal subset of data and local network.*

In recent years, federated learning has paved the way for some promising advances in healthcare by allowing data holding entities to keep their records private while contributing to a large-scale model. FL has been applied to broad range of healthcare data and applications, including disease progression analysis using chest x-rays [16] brain tumor segmentation [11], drug discovery [13], [17], myocardial infarction prediction [18], reflectance confocal microscopy [19], Alzheimer's disease analysis [14], and whole slide image analysis [20]. Toward successful implementation of the FL system, these works have addressed several healthcare data challenges such as security, privacy, interoperability, data integrity, communication efficiency, and heterogeneity [21], [22]. To address the issue of data privacy, [11], [12] used differential privacy mechanisms to improve the privacy of patient data and features. In specific, [11] used selective parameter sharing and differential privacy approaches to provide strong protection against indirect data leakage. While their experimental results showed that there was a tradeoff between model performance and privacy protection costs, their proposed FL procedure achieved a comparable segmentation performance without sharing the client's data.

To improve the security in an FL system, [23] used adaptive federated averaging algorithm to protect sensitive healthcare data from malicious attacks and model poisoning, [24] used masks and homomorphic encryption to prevent the attacker from inferring private medical data through various attacks such as model reconstruction attacks or model inversion attacks, and [22] proposed FRESH, a comprehensive smart healthcare framework for sharing data based on FL and ring signature defense against attacks. To make an FL system scalable, [21] proposes an open-source federated learning frontend framework for healthcare based on a general architecture that allows the integration of new models and optimization methods. [25] proposes SCALT which can be easily scaled or transferred to learn new classes or tasks. To reduce computing overhead while utilizing homomorphic encryption on high-dimensional data, only a variable termed data quality was encrypted for each client in each training epoch [24]. An improved batch verification algorithm is built into [22]'s proposed ring signature schema to leverage the additivity of linear operations on elliptic curves and to assist in reducing the server's computational workload.

Because of differences in demographics, lifestyles, and other aspects, there has been a longstanding goal of achieving a personalized healthcare system. FedAP [9] learns similarities between clients based on the statistics of batch normalization layers while preserving the client's individuality with different local batch normalization to develop personalized models for each client. [26] propose MetaFed, a novel meta-FL framework for healthcare based on cyclic Knowledge Distillation (KD), that can accumulate common information from different federations without compromising privacy and security and achieve personalized models for each federation through adaptive KD. In the 2DFL network [27], heterogeneous data sharing across multiple personal devices and multiple users contributes to the development of a more precise personalized HAR model. Separately, in a FL framework, not all the nodes contribute equally to improving model performance. [28] proposed a contribution-aware FL approach that provides a novel FL aggregation approach to select the best-performing sub-model to be distributed to the FL participants for the next round of local training. [24] considers the local dataset quality of participants to be the main factor in measuring the contribution rate of the local model to the global model in each training epoch. The ensemble model weights in [29] were determined using model accuracy evaluation. This allowed misbehaving or lower-performing models to contribute less to the outcome. In contrast to traditional distributed optimization, FL is faced with statistical heterogeneity i.e., non-identically distributed data across the network [10]. FedProx [10] added a proximal term to generalize and reparametrize FedAvg to improve the stability of the method and to account for data heterogeneity, [30] uses local batch normalization to alleviate the feature shift before averaging models to address non-identical-independent (non-iid) training data. Overall, these approaches addressing several issues concerning healthcare data have shown clear promise of the impact federated learning could have in the healthcare field. However, it is important to point out that the datasets used for these approaches are single modality, and hence the obtained results are suboptimal.

## Multimodal Federated Learning in Healthcare

The current healthcare AI landscape is facing a big shift from narrow, unimodal tasks, confined to clinical data or medical scans to broaden machine capabilities by including other clinical information, including biospecimens and omics data, and encompassing all input modes together [31]. The primary objective in multimodal learning is to integrate diverse data sources to enhance the performance of deep learning models in tasks that require a comprehensive understanding and integration of information across these modalities. This integration is accomplished by concurrently handling data from diverse modalities.

Multimodal AI has shown improved performance compared to its unimodal counterparts across several healthcare domains. In Alzheimer's disease (AD) analysis, Qui et al. [32] combined routinely collected clinical information, including demographics, medical history, neuropsychological testing, neuroimaging, and functional assessments, to accurately classify subjects with normal cognition (NC), mild cognitive impairment, and AD. For cancer survival analysis, MultiSurv [32] used feature representations of clinical, imaging, and different high-dimensional omics data modalities, and aggregated them to generate conditional survival probabilities for follow-up time intervals spanning several decades. Similarly, multimodal deep learning is proposed for cardiovascular risk prediction combining fundus images and traditional risk factors [32]. These representative works clearly demonstrate the benefits of multimodal AI. However, most of these works are developed, trained and validated using large multimodal databases in a centralized system (e.g., UK Biobank [33], the Cancer Genome Atlas (TCGA) [34], etc.), and the multimodal medical database is still heavily restricted, hindering the prospect of multimodal AI in healthcare. Toward this Multimodal Federated Learning (MMFL) could potentially provide an elegant alternative to utilize vast medical information scattered across different hospitals and health centers and build an aggregated general framework to realize the full potential of multimodal artificial intelligence in health. What sets multimodal federated learning apart is its unique ability to amalgamate multimodal information from disparate sources to enhance the robustness and comprehensiveness of medical analyses while preserving patient privacy. This approach distinguishes itself from other subfields by capitalizing on the synergy of multiple data types, fostering a holistic understanding of patient conditions and paving the way for more accurate, personalized, and data-driven healthcare interventions. In our approach, we first conduct a thorough review of MMFL methodologies in healthcare, categorizing the existing literature based on clinical applications. Subsequently, we discuss data partitioning and client management, which is an important part of building accurate and reliable FL systems. We conclude the section with an analysis of the methodological approaches for handling different combinations of modalities at the client level. We highlight the strengths and weaknesses of each approach, along with various technological advances that are brought upon by solving certain problems within the field. This perspective not only highlights the practical implementations of MMFL but also underscores the significant technological progress in this field.

## Current MMFL Applications in Healthcare

*Detection and Diagnosis:* The most common application of MMFL in healthcare relates to the task of disease detection/diagnosis via various combinations of medical scans, electronic health records (EHRs), and sensor data. [35] aims to identify melanoma from skin lesion medical images and clinical data such as gender and malignancy status of the lesion. This work simulates a federated environment by distributing its central data across five client nodes. The system is trained using standard FL approaches and FedAvg model aggregation.

While the aforementioned work focuses primarily on medical data collected by hospital administrators, other approaches aim to utilize IoT data for predictive tasks. [36] implements a personalized model for smart medical care to work toward automated detection of Parkinson’s or cardiovascular disease by merging previously collected CT images with real-time heart rate data collected by personal devices such as smartwatches. In this case, the hospitals and IoT devices communicate directly with one another. The FL system in this work takes on a somewhat inverted structure where there are many servers attached to one client. The hospital (client) transmits encrypted CT image features to the user’s wearable device. This device simultaneously acts as a client and a server, capturing the real-time heart rate data while also maintaining the aggregation module.

Wearable devices do not need to rely on clinically collected data for medical predictions. FitBit data such as caloric intake, steps, and travel distance, along with personal questionnaire information has been utilized by [37] to predict sleep quality in participating adults. This can lead to the early detection of sleep disorders, which contribute to the development of various diseases such as obesity, weakened immune system, high blood pressure, memory loss, stroke, and cardiovascular disorders. In this system, IoT devices communicate directly with the server and train via standard federated learning approaches.

The COVID-19 pandemic prompted swift development of MMFL techniques. During this time, hospitals around the world saw an influx of important medical information regarding the disease but could not adequately share it with ML researchers due to HIPAA and GDPR regulations. As such, federated learning became an attractive solution to this problem. Computed tomography (CT) is the most common modality used across all MMFL models for COVID-19 diagnosis, used in approaches proposed by [38], [39], [40]. In addition to CT scans, chest X-ray [38], EHR, biochemical, personal identity [39], cough audio, PET scan, and RT-PCR [40] data has been utilized to supplement these models. On the other hand, [41] does not use CT scans at all, instead opting to use chest X-rays and ultrasounds for covid discovery.

Particularly in medical imaging tasks, it is often difficult to obtain multimodal pairs of data aligned with a single subject. [38] addressed this issue by adopting a Siamese network to learn image embeddings based on semantic similarity. Siamese networks learn by simultaneously embedding two images and comparing them via a similarity score. This means that the input images do not need to belong to the same patient. If both images correspond to a positive COVID-19 diagnosis, they will result in a high similarity score, otherwise the score will be low. This technique allowed [38] to combine unrelated CT and chest X-ray COVID-19 datasets to train a more powerful multimodal predictive model. Conversely, [41] addresses this problem by allowing client models to remain fully unimodal while parameter aggregation in the cloud handles the multimodal fusion. This process allows the final model to make predictions on both X-ray and ultrasound images despite client models being trained only on one or the other.

*Segmentation:* Recent work as also attempted to improve medical imaging segmentation via federated learning and U-net architecture [38], [42], [43], [44]. [43] utilizes paired magnetic resonance imaging (MRI) and CT scans to perform liver segmentation for the early diagnosis of liver cancer. This model is designed such that hospitals are not required to maintain both MRI and CT facilities, rather each can train their local model with one or both modalities. Details of this technique are further discussed in **Client Modality Management**. Similarly, [42] achieves similar flexibility by training a brain tumor segmentation model to accept T1, T2, T2-FLAIR, and native MRI scans as input. Lasty [44] aims to improve lesion segmentation algorithms by applying FL for CT and PET scan-based kidney segmentation. In this work, the authors simply train two unimodal clients and aggregate the models via FedAvg. Interestingly, this approach does not seem concerned with distributed learning for improved privacy, rather it seeks to exploit the elegant model aggregation strategy employed by FedAvg as a simple modal fusion strategy.

*MRI reconstruction and synthesis:* MRI reconstruction refers to the task of converting raw scan data to a human interpretable image. Here, the modalities are various MRI resonances (i.e., *T1w, T1ce,* and *T2w).* FedMRI [45] employs a cascading U-net model where only the encoder half is shared with the server for global aggregation from frequency and image domains. The decoder block is left to be completely private at the local level. This technique allows the decoder to explore domain-specific properties while the encoder handles general representations of the data. Meanwhile, each client in [46] performs feature disentanglement on their private data to learn modality-specific and modality-invariant representations. The modality-specific features aim to address the problem of domain shift between MRIs of different modalities. The modality-invariant features are sent to the centralized server which utilizes a similarity score to force these features to remain consistent across all clients. This helps exploit useful information that is consistent across all clients. The combination of these two techniques allows for superior reconstruction performance.

MRI synthesis involves computing contrasts such as T1 and T2-weighted images mathematically to accelerate the scanning process. This is desirable as it helps reduce artifacts caused by motion during an otherwise long scan, reducing the likelihood of repeated scans for the same weight and consequently costs. FedMed-GAN [47] employs a GAN-based approach to MRI synthesis from brain scans. A generator is trained to model modality *B* from input *A* (i.e., convert *T1*-weighted scan to a *T2* scan). The discriminator's role in this scenario is to judge whether a given *B* scan was synthesized or came from the set of real images. In this system, only the generator's weights are shared with the server. The discriminators are left entirely local to further increase privacy. Domain shifts due to data heterogeneity often lead to a compromised model across sites, resulting in suboptimal performance [48]. pFLSynth [48] introduces a personalization block to the generator's architecture. The role of the personalization block is to tune the statistics of site- and contrast-specific variables via adaptive instance normalization and weight channeling. This essentially has the same role as the disentanglement module presented by [46].

*Survival Analysis:* Survival analysis is most often associated with estimating the time until death for a patient with a terminal illness but can more broadly be thought of as a time-to-event prediction task. The event prediction could be death, the progression of some disease such as Alzheimer's Dementia, or another task beyond the scope of healthcare. Despite being introduced in 1972, the Cox Proportional Hazards (CPH) [49] model remains one of the most used models for modern survival analysis. Work by Audreux et al. demonstrates that training the CPH model in a federated manner is difficult due to the non-separability of its loss function. As such, they introduced a discrete-time extended stacked Cox model with a separable loss function to address this problem [50]. The authors demonstrate its effectiveness on complex data by pairing tabular data from The Cancer Genome Atlas (TCGA) dataset with their corresponding histopathology slides.

FedSurf++ [51] utilizes Random Survival Forests (RSF) for survival analysis in a federated setting. Each client trains its own RSF model based on their local dataset. After some number of iterations, the central server selects the best performing trees from each client for aggregation based on a variety of different evaluation metrics including Concordance Index (C-Index), Concordance Index with ICPW weighting (C-Index-IPCW), and Cumulative Area-Under-the-Curve (Cumulative AUC). The authors empirically demonstrate that FedSurf++ attains performance on-par or better than neural network approaches with only a single round of communication.

Competing Risk Analysis (CRA) is a subclass of survival analysis that deals specifically with patients who have been diagnosis with two chronic illnesses. For example, a patient diagnosed with cancer and heart disease is simultaneously at risk dying from both. These are considered competing events. CRA aims to analyze a patient's risk in the presence of both diseases. To handle this, [52] proposed first deriving pseudo-values to augment incomplete data in a federated fashion via their Fedora approach. Then, transformer based CRA models, TransPseudo, are deployed at each client and trained with standard federated learning techniques with the augmented data.

*Human Activity Recognition:* Human activity recognition (HAR) is the process of identifying human activities through data. HAR is particularly important for remote patient monitoring, fall detection, and other healthcare tasks. It can be subdivided into generalized and personalized HAR. Generalized human activity recognition aims to identify natural human movements such as sitting, walking, running, climbing stairs, or cycling [53], [54]. Personalized HAR specializes in monitoring health through human activities and physiology [55], [56], [57]. Human activity recognition through federated learning is important in healthcare as it enables possibilities for real-time patient monitoring through IoT devices without compromising privacy.

CMF-HAR [53] distributes video, audio, optical flow, sensor, and image data across a federated system such that each client is unimodal, but the system as a whole is multimodal. This allows each client to focus their individual learning on a single task while contributing to the overarching task of multimodal human activity recognition. In contrast, [54] uniformly distributes multimodal accelerometer, gyroscope, and magnetometer data across the system to achieve the same task of generalized human activity recognition. This work is unique in that each client maintains its own model architecture that is best suited for its private data. FedStack [54] is introduced as an alternative to standard aggregation techniques such as FedAvg to accommodate this system. We further elaborate on this in **Client Modality Management.**

Personal Identification (PI) and Personal Movement Identification (PMI) are terms that are often used to describe personalized HAR tasks. These put more emphasis on identifying human activity for monitoring health. [56] combines federated electrocardiogram (ECG) and radar data for vital signs monitoring. While this achieves the goal of personal identification, there is incentive to utilize Internet-of-Things (IoT) devices such as wearables and in-home camera systems for real time patient monitoring. However, it is not feasible to manually label the high quantities of data that are generated in real time systems. As such, [55] automatically labels these data via a Deep Reinforcement Learning (DRL) component on the client device. Afterwards, the labeled data are classified by using a federated Bi-directional Long Short-Term Memory (Bi-LSTM) module, which is shared with the server. Other work [57] applies a semi-supervised algorithm by supplementing the unlabeled client data with a small, labeled dataset maintained by the server. This solution otherwise maintains standard FL training approaches.

**Data Partitioning and Client Management**

Federated learning paves the way for AI powered healthcare for diagnosis, real-time patient monitoring, survival analysis, etc. However, the data in a real world federated learning environment is naturally heterogeneous, which complicates the training process. This problem is discussed in detail **Challenges in Multimodal Federated Learning.** From a model deployment perspective, an FL framework must meet two primary criteria to successfully replicate a realistic scenario. Results from an unrealistic FL framework are less reliable and should not be deployed for healthcare tasks. We define these criteria as follows:

Firstly, Clients should be representatives of different institutions. In tasks involving data agencies such as hospitals, there should be at least one client in the system that contains modality data that represents a domain shift or non-iid structure. This follows the assumption that different hospitals have access to various quality of medical imaging instruments or see differing patient inflow patterns. For example, a realistic FL environment would have one client with high quality MR Images and another with lower quality ones. For IoT tasks, the clients should be representative of devices of different models and with different sensors. [37] uses a dataset in which all data was collected via FitBit v2. This is acceptable if the model is only deployed on FitBit devices. However, it would likely not perform as well on other wearables since the different sensors would introduce heterogeneity that was not seen during training. A common approach to simulate a realistic environment is to accumulate multiple related datasets and treat each as its own client, rather than partitioning data from a singular dataset [41], [43], [45], [48], [53]. Some datasets on the other hand contain multi-institutional data, which can be partitioned as the client devices to achieve the same effects as the previously mentioned multi-dataset approach [52]. Others merge multiple datasets into one large dataset which is uniformly distributed across the system [38], [55], [58]. While this solution is not ideal, it is useful in situations where there are few available datasets for the task.

Second, clients should be treated as having access to varying levels of resources. Hospitals in rural areas would likely have access to fewer computational resources than ones in major cities. This becomes even more important on IoT devices as the disparity between different devices in the system would be more dramatic. For example, a modern smartphone is vastly more powerful than a smartwatch. In this case, the FL system would need to be devised to consider the restrictions of the smartwatch while avoiding *low resource utilization* [59], underutilizing the smartphone's power. [39] simulates this by providing each client node with one of three GPUs: Tesla V100, RTX 2080Ti, and GT 1030. The authors handled the performance discrepancies by providing each client with a variable amount of data such that no client would be underused or overworked.

While these criteria are not required to build an MMFL system for healthcare tasks, they help ensure that models are robust to heterogeneous data. The inclusion of diverse client populations not only enhances the robustness of the FL system, but also contributes to a fair and representative model as a whole. Additionally, acknowledging the various levels of compute power at the client level serves to promote widespread adoption and collaboration. Without this consideration, certain clients may not be able to participate due to their lack of resources.

**Client Modality Management**

Each client in a federated system has unique data availability and storage requirements. As such, it is difficult to ensure that modalities are uniform across the system. We categorize the method with which FL environments handle this challenge into three primary categories based on how different modalities are distributed across the system. The first assumes every mode is available to each client; we henceforth will refer to this as Client Modal-Complete MMFL (CMC-MMFL). Another approach assumes each client contains only one of the available modes in the environment. Maintaining the perspective of the client, we call this Client Unimodal MMFL (CU-MMFL). Finally, we must account for the case where any individual client may contain any combination of modalities, which we refer to as Client Modal-Incomplete MMFL (CMI-MMFL).

We elaborate further on these conventions below and provide formal definitions. For the following subsections, let $C$ be the set of all clients in a Multimodal Federated Learning environment and $c_i$ as client $i$ in $C$. Additionally, we define the combined dataset of all clients in $C$ as $D = \{c_1, c_2, \dots, c_n\}$, where $n = |C|$. Further, allow $M = \{m_1, m_2, \dots, m_k\}$ to be the set of all modalities available across all clients, where $k$ is the total number of unique modalities. Finally, $c_i^{m_j}$ will correspond to modality $m_j$ for client $c_i$.

Additionally, Table 1 categorizes each work based on task, modality distribution, and generalization status. Here, "generalization status" refers to whether the application is tailored toward generalized healthcare as is the case with applications such as segmentation and MRI reconstruction, or personalized healthcare, such as HAR.

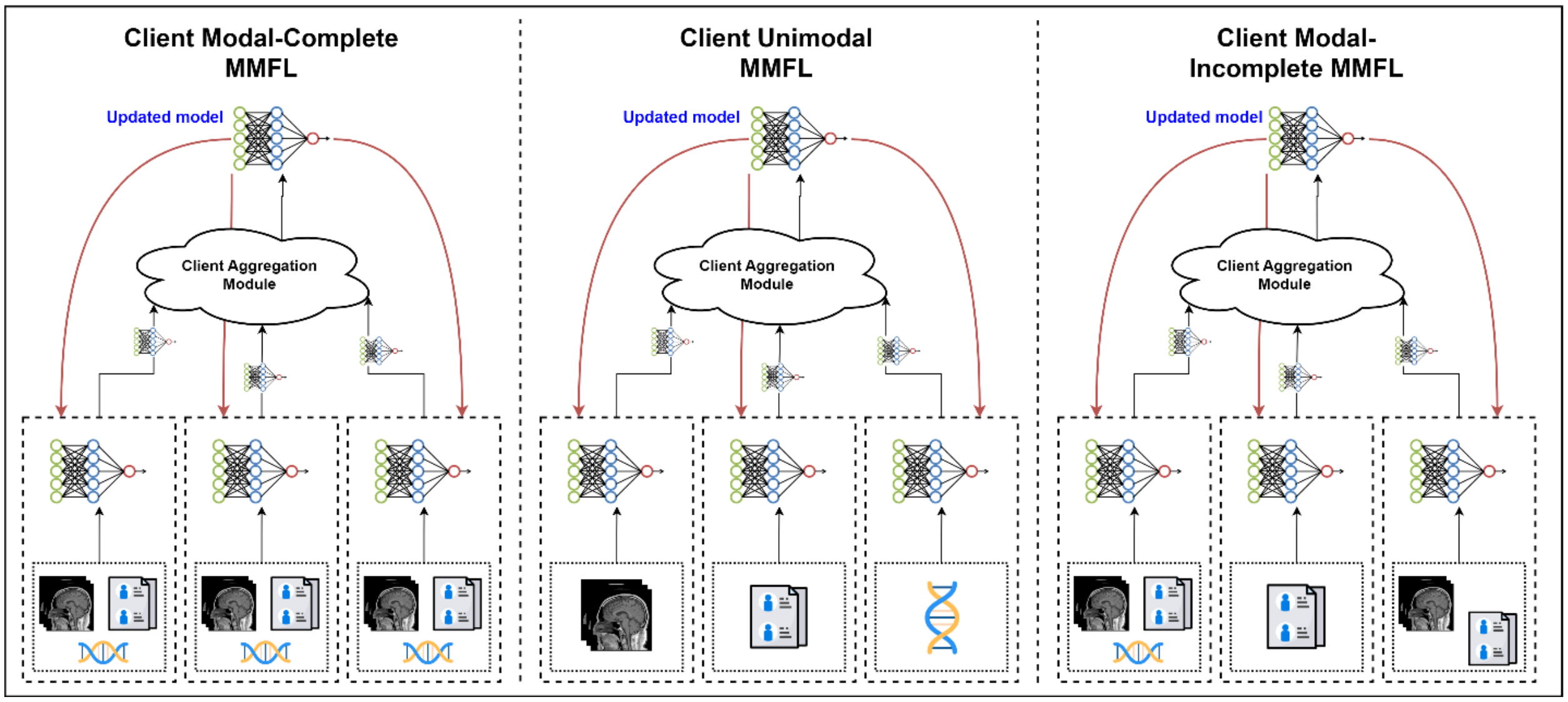

***Figure 3 (left)*** *Modal distribution in Client Modal-Complete MMFL environment. Each client maintains all available modalities (e.g., medical scans, clinical notes, omics data, etc.) in the total dataset D.* ***(Middle)*** *Client Unimodal MMFL system, whereby each one of a subject's modal information is held by a different client, which contains a maximum of one mode.* ***(Right)*** *Client Modal-Incomplete MMFL model, where each client can maintain any combination of the available modes in the Multimodal Federated Learning environment.*

*Client Modal-Complete MMFL:* We formally define a CMC-MMFL system as:

$$D_{c_i} = \{c_i^{m_1}, c_i^{m_2}, \dots, c_i^{m_k}\}, \forall c_i \in D \tag{1}$$

by asserting that each client in $C$ contains information from every mode available in dataset $D$. Intuitively, we can think about each client in the system as having identical data structures as demonstrated by Figure 3 (left), but different subject distributions. These approaches have the advantage of simplifying the FL framework by assuming each model is uniformly structured, but often do not truly represent a realistic scenario. This may be reasonable in applications such as the melanoma classification task presented by [35] as hospitals are likely to simultaneously record image and EHR data for patients with skin lesions. CMC-MMFL could also be a viable option for MRI reconstruction and synthesis [42], [45], [47] tasks as the modalities utilized in these tasks are typically MRIs of varying weights, which a hospital is likely to collect. On the other hand, it can be unrealistic in scenarios where clients potentially have varying sensor availability. The work by [55] on personal movement identification via wearable smart devices may struggle in real world applications as it assumes all devices to maintain the same suite of sensors, which include sound level, GPS, magnetic field, and acceleration data. A device without any one of these facilities would not be able to participate in the FL system because of data mismatch.

*Client Unimodal MMFL:* Conversely, a CU-MMFL environment is defined as:

$$D_{c_i} = c_i^m, m \in M, \forall c_i \in D \tag{2}$$

where every client contains only one mode from *M*. It is assumed that multiple clients contain different modal information for the same subject, visualized in Figure 3 (middle). For example, hospital A could maintain medical imaging records from MRIs or CT scans, while hospital B collects a set of clinical notes for the same subject, all of which are aggregated in the cloud for prediction.

Like CMC-MMFL, this system has the advantage of assuming the clients are uniformly formatted but comes with its own set of challenges. Firstly, multimodal fusion cannot be done at the client level because all modes belonging to a selected subject are fragmented across multiple client nodes. Due to the nature of FL systems, multimodal fusion in this instance must be done in the cloud. It is common for CU-MMFL systems to simply utilize FedAvg to simultaneously aggregate client models and perform multimodal fusion [41], [44]. While this solution is very neat, it also forces each client to maintain identical models, otherwise FedAvg fails. In the circumstance from the example above, client A holds medical images and client B, EHR data. These two modalities require very different client-side models, which would not be compatible for simple weight averaging in the server.

CU-MMFL works well for scenarios such as [46] where the modalities are simply MRIs of different weights (T1w, T1ce, and T2w). Here, the modality invariant features are extracted in each client and passed to the server. This effectively creates a unimodal task from the server's perspective. Rather than fusing multiple modalities, the server simply backpropagates a consistency loss to ensure the modality-invariant features are similar across all clients.

FedStack [54] provides a solution to the model aggregation problem. Each client maintains a unique model that works best for its private data. This could be as simple as a linear regression model or as complex as a transformer. The local models train on their respective client data, as is typical in a federated environment, but rather than transferring the model weights the server, FedStack proposes stacking the model predictions as the aggregation technique. This provides complete freedom for the clients to build their system in a way that makes the most effective use of their local data, if the prediction classes are uniform across the FL system.

*Client Modal-Incomplete MMFL:* Finally, we formalize CMI-MMFL as:

$$D_{c_i} = \{c_i^m \mid m \in M\}, \forall c_i \in D \tag{3}$$

where each client can maintain any combination of modalities, demonstrated by Figure 3 (right). This is the most realistic scenario as it allows clients to maintain any combination of modalities. [43] implements Mode Normalization (MN) [60], which attempts to classify the mode of each sample in a minibatch before applying normalization. This helps the neural network differentiate between the latent space belonging to different modes. With this knowledge, each mode can be treated differently, rather than applying the same statistics to every item in the batch. Bernecker et al. use this technique to introduce FedNorm+ [43] to perform liver segmentation from MRI and CT scan data.

In Fed-PMG [61], each client contains any combination of *T1w* and *T2w* MRI images. Prior to training, the amplitude and phase spectrums for all images from modal-complete clients are collected via multidimensional Fourier Transform. The amplitude spectrums are then sent to the server. Privacy is maintained in this instance because the original image cannot be reconstructed via inverse Fourier Transform without the phase spectrum information. To address the modality discrepancies, the authors introduce a Pseudo Modality Generation (PMG) [61] method to create pseudo modes on the unimodal clients. PMG uses the amplitude spectrums that were sent to the server to generate pseudo data to be used in place of the missing modality. This simulates a CMC-MMFL environment, allowing the system to be more easily trained with standard approaches.

Conversely, Zhao et al. introduce Multimodal FedAvg (Mm-FedAvg) [57] to extend the standard FedAvg algorithm to simultaneously aggregate the unimodal and multimodal models. Each multimodal client maintains two autoencoders, one for each modality, while the unimodal clients simply have one. The global model is a Deep Canonically Correlated Autoencoders (DCCAE) [62] which aims to maximize the canonical correlations between the two hidden representations. In this way, the overall system is flexible to varying modalities.

**Table 1. MMFL applications in healthcare**

| Title | Task | Modality Distribution | Personalized /Generalized | Ref |
|---|---|---|---|---|
| Multimodal Melanoma Detection with Federated Learning | Detection/Diagnosis | CMC-MMFL | Generalized | [35] |
| FL-PMI: Federated Learning-Based Person Movement Identification through Wearable Devices in Smart Healthcare Systems | HAR | CMC-MMFL | Personalized | [55] |
| A Collaborative Multimodal Learning-Based Framework for COVID-19 Diagnosis | Detection/Diagnosis | CMC-MMFL | Generalized | [38] |
| Auxiliary Diagnosis of COVID-19 Based on 5G-Enabled Federated Learning | Detection/Diagnosis | CMC-MMFL | Generalized | [39] |
| A Novel Partitioning Approach for Multimodal | Segmentation | CMC-MMFL | Generalized | [42] |
| Brain Tumor Segmentation for Federated Learning | | | | |
| Specificity-Preserving Federated Learning for MR Image Reconstruction | MRI Reconstruction | CMC-MMFL | Generalized | [45] |
| FedMed-GAN: Federated Domain Translation on Unsupervised Cross-Modality Brain Image Synthesis | MRI Synthesis | CMC-MMFL | Generalized | [47] |
| Federated Survival Analysis with Discrete Time Cox Models | Survival Analysis | CMC-MMFL | Generalized | [50] |
| Scaling Survival Analysis in Healthcare with Federated Survival Forests: A Comparative Study on Heart Failure and Breast Cancer Genomics | Survival Analysis | CMC-MMFL | Generalized | [51] |
| Enhancing Privacy-Preserving Personal Identification Through Federated Learning with Multimodal Vital Signs Data | HAR | CMC-MMFL | Personalized | [56] |
| Collaborative Federated Learning for Healthcare: Multi-Modal COVID-19 Diagnosis at Edge | Detection/Diagnosis | CU-MMFL | Generalized | [41] |
| Cross Domain Federated Learning in Medical Imaging | Segmentation | CU-MMFL | Generalized | [44] |
| Personnel status detection model suitable for vertical federated learning structure | Detection/Diagnosis | CU-MMFL | Personalized | [36] |
| Cross-Modal Vertical Federated Learning for MRI Reconstruction | MRI Reconstruction | CU-MMFL | Generalized | [46] |
| Cross-Modal Federated Human Activity Recognition via Modality-Agnostic and Modality-Specific Representation Learning | HAR | CU-MMFL | Generalized | [53] |
| Multimodal Federated Learning on IoT data | HAR | CMI-MMFL | Personalized | [57] |
| FedNorm: Modality-Based Normalization in Federated Learning for Multi-Modal Liver Segmentation | Segmentation | CMI-MMFL | Generalized | [43] |
| Federated Pseudo Modality Generation for Incomplete Multi-Modal MRI Reconstruction | MRI Reconstruction | CMI-MMFL | Generalized | [61] |
| One Model to Unite Them All: Personalized Federated Learning of Multi-Contrast MRI Synthesis | MRI Synthesis | CMI-MMFL | Generalized | [48] |
| FedStack: Personalized Activity Monitoring using Stacked Federated Learning | HAR | CMI-MMFL | Generalized | [54] |

## Challenges in Multimodal Federated Learning

Despite these advances, the field of multimodal federated learning faces major technical challenges, particularly in dealing with heterogeneity related to data and resources. Understandably, extending federated learning to the multimodal domain exacerbates already formidable issues present with unimodal data. Further, the nature of healthcare naturally lends itself to encountering all sources of heterogeneity simultaneously when applied to a federated framework. Figure 4 presents an overview of the heterogeneity challenges brought upon my extending MMFL to healthcare. There also exist complications relating to general AI in health, including addressing bias and fairness, transparency, and interpretability, which must be considered on top of the challenges from FL.

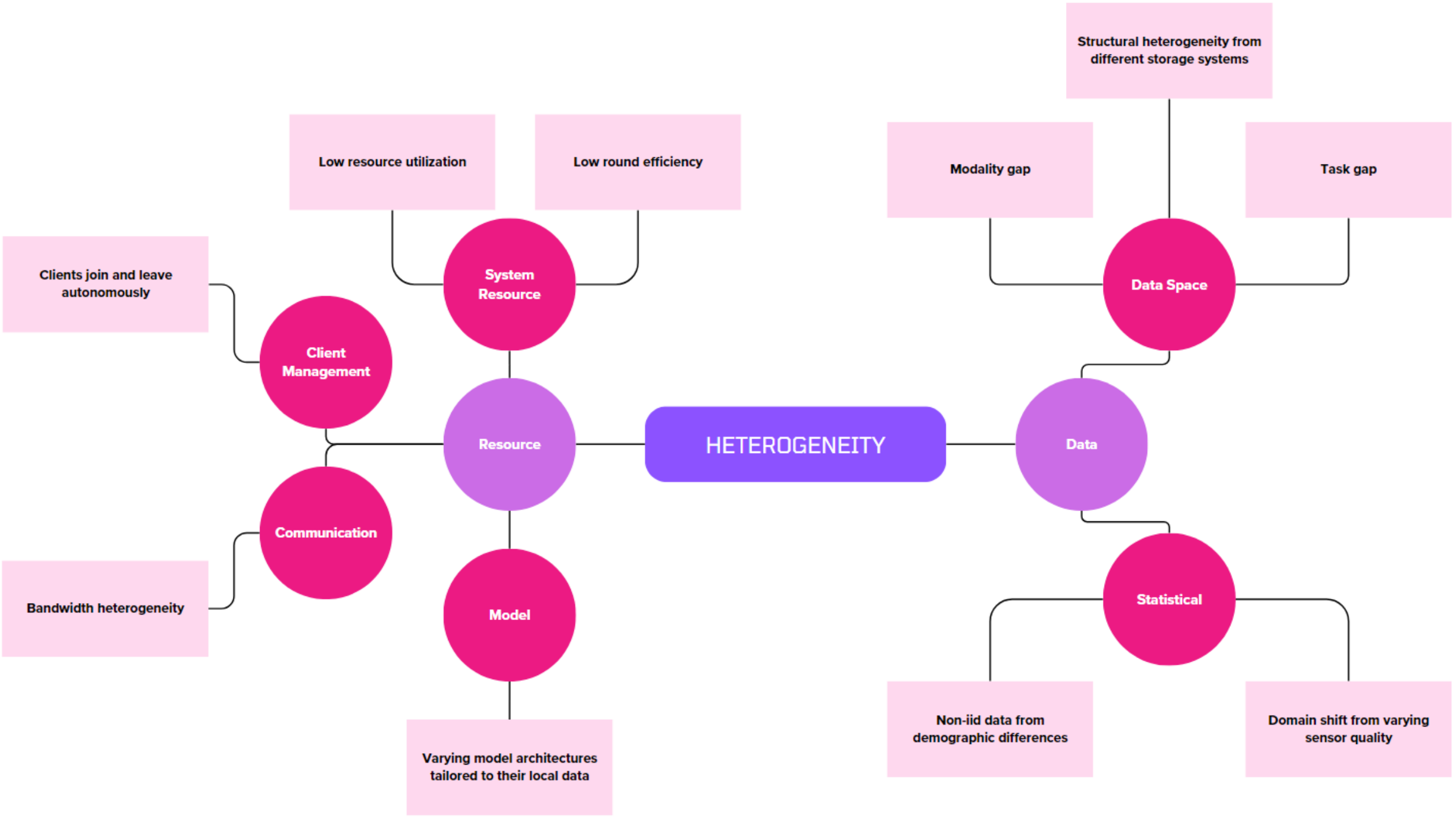

***Figure 4*** *Overview of heterogeneity sources caused by MMFL in healthcare.*

**Data Heterogeneity**

Data heterogeneity is a result of the diversity in data collected from different sources such as varying data formats, feature distributions, or noise levels. This represents one of the most pressing issues in federated learning as it significantly impacts training time and reduces accuracy in the MMFL pipeline [63]. Data heterogeneity can be further divided into data space and statistical heterogeneity [59].

*Data Space Heterogeneity:* This refers to the difference in the feature space or label space of the data collected from different sources. In healthcare, data can be collected from a wide array of sources such as Electronic Health Records (EHRs), IoT devices, and clinical trial examinations. These data all reside in different feature spaces, which each have unique properties. Furthermore, the same data can be collected from different sources such as hospitals, clinics, research institutions, or wearable devices. If there is no universal standard for handling such data, each client may store it differently. There are three overarching storage methods for this: structured, semi-structured, or unstructured. Structured data is highly organized in a digestible database, while unstructured data has no defined system. Each of these systems must be considered when implementing a federated learning environment.

Incorporating multiple modalities into the federated system further exacerbates the data space heterogeneity problem [64], [65], [66]. Each client in the system maintains its own set of devices, be that medical imaging equipment in hospitals, sensor data in IoT devices, or some other data collection mechanism. As such, it is very likely that not all clients in the system will share the same facilities. This is a common occurrence in real scenarios due to constraints in hardware, data collection errors, or storage limitations. For instance, one client may have both CT and MRI image samples whereas another client may have only CT image samples. Consequently, clients must train in different tasks because they do not have an identical set of input modalities. For example, an Image-Text client may be trained on retrieval tasks whereas an Image-only client could be trained on classification tasks. Therefore, this setup leads to distinct parameter space among clients which increases the difference in local objective functions. Hence, the one-model-fits-all formulation of federated learning [1], [10] proves inadequate.

This modality heterogeneity is directly related to the modality and task gap [67]. The modality gap indicates differences in representation and semantics among various data modalities. For instance, an image of a person might have different representation and semantics compared to a text description of the same person. Modality gaps can impact performance and fairness [68]. On the other hand, the task gap refers to variations in training objectives among clients. For instance, a unimodal client might be training a model to classify images of healthy and unhealthy medical scans, whereas a multimodal client might be training a model to classify both medical scans and clinical textual descriptions of healthy and unhealthy subjects. Additionally, the challenge of multimodal alignment exists, involving finding associations between different modal instances. This alignment task necessitates aligning data from diverse modalities without explicit labels, posing limitations for real-time performance and reducing training time [65].

Some early attempts to address modality heterogeneity in MMFL include MMFed [69], and FedIOT [57]. MMFed [69] introduces FedAvg in multi-modal scenarios. Similarly, FedIOT [57] also employs the FedAvg scheme in a multi-modal federated setting, where parameters from local split autoencoders trained on different modalities are aggregated. The global autoencoder is subsequently updated through the aggregation of both unimodal and multi-modal clients, whereby the multi-modal clients are assigned a greater weight in the aggregation process. MMFed does not take into consideration any of these problems whereas FedIOT does take into account missing modalities, but their method is ineffective for heterogeneity caused by limited client resources.

Yu, Qiying, et al [66]. propose Contrastive Representation Ensemble and Aggregation for Multimodal FL (CreamFL) to address the challenges posed by task heterogeneity and missing modalities in heterogeneous architectures. They employ contrastive-based regularization in the training of clients to mitigate task differences and missing modalities. Furthermore, they incorporate knowledge distillation in the form of feature distillation, where they aggregate the features of clients using a global-local cross-modal contrast approach. The authors leverage a publicly available dataset for knowledge distillation and contrastive-based regularization. However, to achieve effective knowledge distillation, a significant amount of public data samples are required, which consequently leads to a significant increase in communication costs between clients and server communication which causes problems in limited wireless bandwidth scenarios. Similarly, Le, Huy Q., et al. [70] introduce FedMKT which allows the server and clients to exchange knowledge about the data without sharing the data itself through the utilization of a small proxy dataset. However, their knowledge transfer scheme is heavily based on proxy datasets and the quality of the proxy datasets determines the effectiveness of knowledge transfer schemes.

While not originally introduced to address the modality gap, FedStack [54] offers an elegant solution to the problem. Rather than passing the model parameters to the server, which requires all client models to have the same architecture, FedStack instead transmits only the model predictions. As long as each client maintains the same task, this allows every client to have a unique architecture that fits their needs. In this way, a medical imaging client and an EHR-only client can easyily fit within the same federated learning framework without the need to align their models.

*Statistical Heterogeneity:* Statistical heterogeneity refers to the statistical variation of data collected from the same data space, referred to as non-iid (independent and identical distributions) [59]. This can happen for a variety of reasons, such as different environmental conditions, different user preferences, or different data collection methods. For example, inpatient flow patterns can vary between hospitals, affecting the distribution of patient data across different departments and wards within each institution [71]. The non-IID data distribution among participating devices makes it difficult for the clients to converge to the global optimum [72]. This is because the clients may overfit their local objectives, which can hinder the performance of the central aggregated model [73], [74]. This is already problematic in a unimodal FL environment, but as with data space heterogeneity, the problem worsens when more modalities are added.

The disparities in data modalities among various entities involved in a federated system can be understood through the concept of domain shift [75]. In a unimodal context, domain shift occurs when two clients have different data distributions for the same object. For instance, in the realm of medical imaging, a hospital in a major metropolitan area might have high-end X-ray equipment, while a hospital in a rural area might have a lower quality one due to limited resources. Both sets of X-rays offer valuable information for training, but they might exhibit different histogram distributions, complicating the process of data alignment. This challenge is further amplified in the presence of multimodal data, as alignment between different modalities must also be considered.

Yan et. al. [46] attempt to address domain shift between MRI scans of varying intensities (T1w, T2w, and T1ce) in a CU-MMFL approach by simultaneously maintaining two datasets. They define $\mathcal{D}_k^H$ as the horizontal dataset for client *k*, which represents MRIs from non-overlapping subjects and $\mathcal{D}_k^V$ as the vertical dataset for MRIs belonging to subjects that overlap across multiple clients. This allows intra-client feature disentanglement to extract features for MRI reconstruction while utilizing $\mathcal{D}_k^V$ to act as a domain anchor to maintain latent representation consistency across clients. Meanwhile, [43] utilizes Mode Normalization to classify the type of data for each sample in minibatch in order to differentiate the latent space between modes. This allows the model to treat each mode differently.

Although several strategies have been explored to address these challenges, including personalized federated learning variations [76], [77], [78], the application of data augmentation techniques [79], the use of distributional transformations [80], and the pursuit of virtual homogeneity learning approaches [81], more principal efforts are required for successful realization of multimodal federated learning.

**Resource Heterogeneity**

Data is not the only source of heterogeneity in federated systems. As each client is fully autonomous, they each have varying levels of resources beyond the previously discussed data collection mechanisms. Hospitals of different sizes and in different cities may have varying amounts of compute power or communication bandwidth. IoT devices potentially suffer even more as considerations should be made for different device models, personal communication resources likely being significantly less than what is available at a hospital, and personal decisions on data privacy, etc must be made.

*System/Resource Heterogeneity:* In federated learning systems, client devices may differ in resources such as availability of graphical processing units (GPUs), storage memory, and network connectivity. This results in varying compute and communication capacity across the environment. A federated learning environment is only as strong as its weakest link. For example, *low round efficiency* refers to the scenario in which clients with faster devices must wait for slower clients in each communication round before aggregation can be performed [59]. Similarly, the problem of *low resource utilization* is defined by competent clients may be underutilized as they scale back power requirements to match the rest of the models [59]. In the healthcare domain, this diversity is exemplified by the coexistence of wearables, home devices, and hospital equipment, each with distinct resource profiles.

*Model Heterogeneity:* One of the challenges associated with Multimodal Federated Learning (MMFL) is the heterogeneity among participating clients in terms of data, resources, and requirements. This heterogeneity necessitates the training of different models to best utilize the available data and resources. For instance, in a healthcare context, one hospital might possess medical image data, while another might only have tabular medical record data due to a lack of imaging devices. Consequently, these two hospitals would need to train different models tailored to their specific data types [59]. Similarly, in the context of smartphones and wearables, wearables typically have limited computing power compared to smartphones. Therefore, smartphones can accommodate larger, more resource-intensive models compared to wearables. However, when these diverse client models are aggregated, the heterogeneity often leads to challenges in convergence [82].

*Communication*: Exchanging training parameters, such as weights and gradients, between clients and the server is essential for MMFL. However, MMFL faces various challenges during this exchange [83]. The communication cost in MMFL is directly proportional to the communication bandwidth required to transfer model parameters between clients and the server. It is important to conserve bandwidth as millions of training parameters are changed via this communication channel, as not all clients may always have equal bandwidth available. Such clients can create a bottleneck, especially clients such as portable or multi-purpose devices that have lower bandwidth [84]. Low bandwidth can lead to intermittent clients in federated learning systems impacting training by requiring data redistribution and causing delays when clients disappear or join. Moreover, delays in sending or receiving weights for aggregation hampers the training process [85]. To address these challenges, reducing the number of transmitted parameters, compressing the model, sparsification, and quantization [5], [86], and selecting a subset of communication-efficient clients aim to balance communication cost and bandwidth usage.

*Client Management:* Clients in the federated learning system are heterogenous, independent, and autonomous, but this can create several management challenges, especially during the multimodal scenario. Since clients are autonomous, they can join or leave the training process anytime, this hampers the aggregation process and preventing clients from dropping out has been a persistent problem. Clients' heterogeneity in data, models, resources, and availability poses challenges in selecting those contributing valuable data or models, impacting FL reliability [87]. Some clients may only participate when not performing their primary task, necessitating prioritization based on statistical utility and scheduling system utility [88]. Evaluating contributions considers data quality and resource usage [89]. Shapley value, Stackelberg game theory, and auction-based mechanisms are proposed to assess and address such contributions. However, reliable assessment tools specific to multimodal systems are lacking. Separately, to encourage meaningful participation, mechanisms like FedMCCS [90] and Blockchains [91] have been suggested. Client management will not be complete if we don't think from the client's perspective about their issues such as personalization, privacy management, incentive management, resource management, data and device security, and fairness [92].

**Bias**

Extensive research has revealed that AI systems can systematically and unfairly exhibit biases against certain populations in various scenarios, leading to disparate performance for different sub-groups defined by protected attributes such as age, race/ethnicity, sex or gender, socioeconomic status, among others [93]. Addressing this pervasive issue in healthcare and ensuring the development of fair machine learning models is equally crucial for MMFL systems. However, due to the variations in data modality and models, these challenges become even more complex within the context of multimodal federated learning.

*Data Bias:* MMFL integrates data from diverse sources, including medical images, health records, and omics data. Each modality can introduce inherent biases, which may stem from flawed measurements, subjective decisions, or errors. The combination of these modalities can result in complex representation bias. Additionally, missing data and selection bias further contribute by creating non-representative datasets. Moreover, in federated learning, data is decentralized. Variations in data collection practices, patient demographics due to socioeconomic factors, and differences in healthcare settings can introduce data source bias. The heterogeneity of the data can lead to bias since the global model may not accurately represent all groups involved, and this effect is accentuated when participating devices are intermittent.

*Model Bias:* Algorithmic bias in machine learning stems from various sources. The use of proxy variables in place of protected attributes like race or gender can introduce bias when algorithms make decisions based on these proxies [59]. Fusion algorithms, crucial in aggregating local model updates, can introduce bias by favoring parties with larger datasets. Similarly, client selection processes can bias the global model if skewed towards faster devices or specific geographic regions, thereby influencing corresponding user groups [60]. Training a model to perform equally on data from different groups may lead to overfitting on underrepresented groups with less training data, increasing the risk of privacy violations. Conversely, inconsistent reductions in accuracy caused by privacy mechanisms on classification and generative tasks can impact underrepresented groups more than other groups [61].

*Ethics:* A MMFL approach presents several ethical dilemmas. Aggregating data from diverse sources means that inherent biases in individual datasets can amplify within the federated model. Federated learning's decentralized nature can make it challenging to interpret model decisions. Issues of representation and fairness in diagnoses, treatments, and outcomes must be addressed, given that transparency is crucial for patients to make informed decisions. Collaboration and data sharing among institutions demand robust governance frameworks. The equitable distribution of benefits, responsibilities, and data access must be ensured to maintain ethical integrity.

**Other Challenges**

*Security & Privacy:* The major security concerns in MMFL systems involve confidentiality, integrity, and availability, while the main privacy concerns include traceability, identifiability, profiling, and localization [94]. There are various security and privacy attacks that can compromise MMFL, including poisoning attacks, inference attacks, and watermark attacks, and defending against these is critical for MMFL systems. Poisoning attack in multimodal federated learning refers to a type of attack where a malicious actor intentionally manipulates the data or model parameters to corrupt the global model for its accuracy [95]. Data and model poisoning are two major methods of poisoning the MMFL system. Data poisoning occurs when attackers inject malicious data for a single modality or multiple modalities, manipulating the data distribution of the training dataset. This manipulation results in good overall performance but degrades the model's performance for specific victims. Model poisoning attacks involve manipulating training parameters for model updates or inserting a backdoor. Sometimes, both model and data attacks can be applied together [89], [95].

Inference attacks involve attempts by attackers to infer sensitive information about a user's data from the shared model parameters. In MMFL systems, inference attacks pose serious threats to user privacy. There are mainly three types of inference attacks: membership inference attack, property inference attack, and model inversion attack. Membership inference attacks determine if a data point is seen by the model or not. Attackers can observe the global model update passively without affecting the training, or they can inject an adversarial update during training. In a property inference attack, an attacker tries to extract properties of other clients' data unrelated to the model's objectives. Model inversion attacks attempt to recover training data or sensitive attributes from a trained model using the model's gradients, compromising data privacy [96]. Lastly, a watermark attack injects a watermark into the victim's dataset. Attackers then extract sensitive information by comparing the exposure rates of the watermark to the exposure rates of potential interest records. This type of attack is invisible to the victim and does not affect the overall performance of the global model. Overall, building security measures to protect multiple modalities of data and overall MMFL system against these attacks are crucial for their use in real world settings.

*Transparency and Interpretability:* Interpreting MMFL is challenging mainly due to the invisibility of local data due to data privacy concerns [97]. MMFL systems can be vulnerable to attacks such as gradient leakage attacks, where training data can be reconstructed [98], to mitigate, differential privacy, which injects noise in model parameters, is used, however, doing so hinders gradient-based interpretation. In addition, Multimodal Learning inherits its share of interpretability issues [99]. Although transparency and interpretability are vital in healthcare [91], exploration of MMFL interpretability in the healthcare sector is limited [100] such issues make addressing AI interpretation to lay stakeholders even more challenging in MMFL.

## Conclusion

The exploration of multimodal federated learning in healthcare represents a significant stride toward enhancing the efficiency, privacy, and accuracy of healthcare systems. By consolidating diverse data types and harnessing the collaborative power of federated learning, healthcare practitioners and researchers are empowered to extract deeper insights and develop more precise models for diagnosis, treatment, and disease prevention. The reviewed studies underscore the potential of this approach, demonstrating its ability to address the challenges of data silos and privacy concerns prevalent in healthcare data management. However, it is imperative to acknowledge the existing hurdles, including ethical considerations, and technical complexities, that demand continual attention and innovative solutions. As the field of multimodal federated learning evolves, there is a pressing need for standardized protocols, robust security frameworks, and interdisciplinary collaboration among data scientists, clinicians, and policymakers to ensure the responsible and effective implementation of this transformative approach in healthcare.

## Funding

This work is supported by WVHEPC (RCG23-007) ad CSF CiTER (23F-02W)

## Authors Contributions

Conceptualization, J.T., A.D., and P.G.; writing – original draft, J.T., A.D., and P.G.; writing – review & editing, J.T., A.D., P.S., R.C., P.P., C.H., B.B., and P.G.

## Declaration of Interests

The authors declare that they have no known competing financial interests or personal relationships that could have appeared to influence the work reported in this paper.

**Inclusion and Diversity**
We support inclusive, diverse, and equitable conduct of research.